# Attention2Minority: A salient instance inference-based multiple instance learning for classifying small lesions in whole slide images

Ziyu Su*, Mostafa Rezapour, Usama Sajjad, Metin Nafi Gurcan, Muhammad Khalid Khan Niazi
Center for Biomedical Informatics, Wake Forest University School of Medicine, Winston-Salem 27104, USA
*Corresponding author. E-mail: zsu@wakehealth.edu

**Abstract:**
Multiple instance learning (MIL) models have achieved remarkable success in analyzing whole slide images (WSIs) for disease classification problems. However, with regard to giga-pixel WSI classification problems, current MIL models are often incapable of differentiating a WSI with extremely small tumor lesions. This minute tumor-to-normal area ratio in a MIL bag inhibits the attention mechanism from properly weighting the areas corresponding to minor tumor lesions. To overcome this challenge, we propose salient instance inference MIL (SiiMIL), a weakly-supervised MIL model for WSI classification. We introduce a novel representation learning for histopathology images to identify representative normal keys. These keys facilitate the selection of salient instances within WSIs, forming bags with high tumor-to-normal ratios. Finally, an attention mechanism is employed for slide-level classification based on formed bags. Our results show that salient instance inference can improve the tumor-to-normal area ratio in the tumor WSIs. As a result, SiiMIL achieves 0.9225 AUC and 0.7551 recall on the Camelyon16 dataset, which outperforms the existing MIL models. In addition, SiiMIL can generate tumor-sensitive attention heatmaps that is more interpretable to pathologists than the widely used attention-based MIL method. Our experiments imply that SiiMIL can accurately identify tumor instances, which could only take up less than 1% of a WSI, so that the ratio of tumor to normal instances within a bag can increase by two to four times.

**Keywords:** Deep Learning, Whole Slide Image Analysis, Multiple Instance Learning, Weakly Supervised Classification.

___

## 1. Introduction

Histopathology plays a crucial role in clinical diagnosis and is widely used as a diagnostic tool for multiple diseases [1]. In addition, with the fast development in digital scanning, data storage and data transmission technologies, whole slide image (WSI) technology has made histopathological slides more accessible and easier to analyze [2]. WSIs usually have gigapixel-level resolution that enables pathologists to see details of the biological tissues in high magnification. This sheer size of WSI, on the other hand, makes manual delineation and diagnosis/prognosis difficult and time-consuming. Especially, when it comes to micro-tumor metastasis, pathologists could not find clear evidence from the routine hematoxylin & eosin (H&E) WSIs. Thus, pathologists have to create adjacent immunohistochemistry (IHC) staining slides for clarification, which is time-consuming and of low specificity. As a result, there is a growing interest in developing automated methods for mining diagnostic and predictive information from routine H&E WSIs.

Deep learning is a supervised-learning paradigm in which the term "deep" refers to a neural network's numerous hidden layers. Deep learning-based medical image analysis techniques have advanced significantly during the last decade [3]. As a result, deep learning is the common choice for the automatic WSI analysis, showing substantial advancements over earlier approaches [4-7]. However, fully supervised-deep learning models cannot be used for many pathology problems since they require laborious and time-consuming fine-grained annotations. As a result, weakly supervised learning, a paradigm in which the model just needs to be trained using WSI-level diagnostic labels, has risen in prominence in recent years [8-10].

Multiple instance learning (MIL) is one of the most extensively used weakly supervised learning methods for medical images [11]. In MIL, high-resolution medical images are divided into smaller image patches, then the information extracted from these patches is combined to make image classifications. If any patch within the image contains disease-related lesions, the entire image will be classified as unhealthy; otherwise, it will be deemed healthy. Notably, the training process of MIL is completely supervised by diagnostic labels assigned at the image level, without requiring specific annotations for individual patches. MIL inherently aligns with the task of predicting a medical image as having a specific disease or not [11]. This is because disease-related lesions are typically localized within local regions of medical images, while corresponding labels are provided at an image level. MIL has found widespread application in medical image analysis, including tasks such as pneumothorax recognition in chest X-ray images [12, 13], COVID-19 detection in CT images [14-16], retinopathy detection in retinal images [17, 18], and tumor slide identification for WSIs [8, 19, 20]. Especially, WSI is currently the most significant application area for MILs. This is due to MIL's capacity to accommodate ambiguous patch-level annotations, handle large datasets efficiently, and facilitate scalable classification processes for giga-pixel WSIs.

The preliminary study of predicting WSI's slide-level outcome using MIL was done by Courtiol et al. in 2018 [19], in which they identified tumor metastasis in sentinel lymph nodes. They employed MIL model to predict the presence or absence of tumor patches within a WSI. Some other MIL models use neural networks to learn attention weights for each patch of the WSI. These weights can be visualized to learn which parts of the WSI are prioritized by the MIL model when predicting existence of tumor lesions [9, 10, 21, 22]. However, for early diagnosis, the WSI-level labels frequently correspond to exceedingly small lesions (limited to a few patches) in the WSIs, which sometimes account for a very small (typically less than 1%) portion of the total WSIs. As patch data contains a limited amount of meaningful information (patches corresponding to the small lesion), the MIL models fail to pay attention to the small lesion during the analysis. This effect is illustrated in Fig. 1a, b and c, where we overlaid the MIL model [21] generated attention weights over a tumor WSI. Other MIL models attempt to only feed partially important regions of the WSIs to the deep learning models to highlight the important regions of the WSIs [8, 19, 23, 24]. In these models, the essential idea is to train a patch selection network based only on the WSI-level labels and then select a few patches as the representation of the WSI. This kind of methodology works better in situations where the WSIs have predominant regions that correspond to the WSI label, such as cancer subtyping, Gleason grading, and tissue regeneration grading problems [24]. However, these approaches may fail for WSI with small lesions since most of the patches do not correspond to a slide label. Expecting a patch selection network to uncover these minority patches is akin to looking for a needle in a haystack. Given these difficulties, a robust patch selection technique for the MIL models is critical.

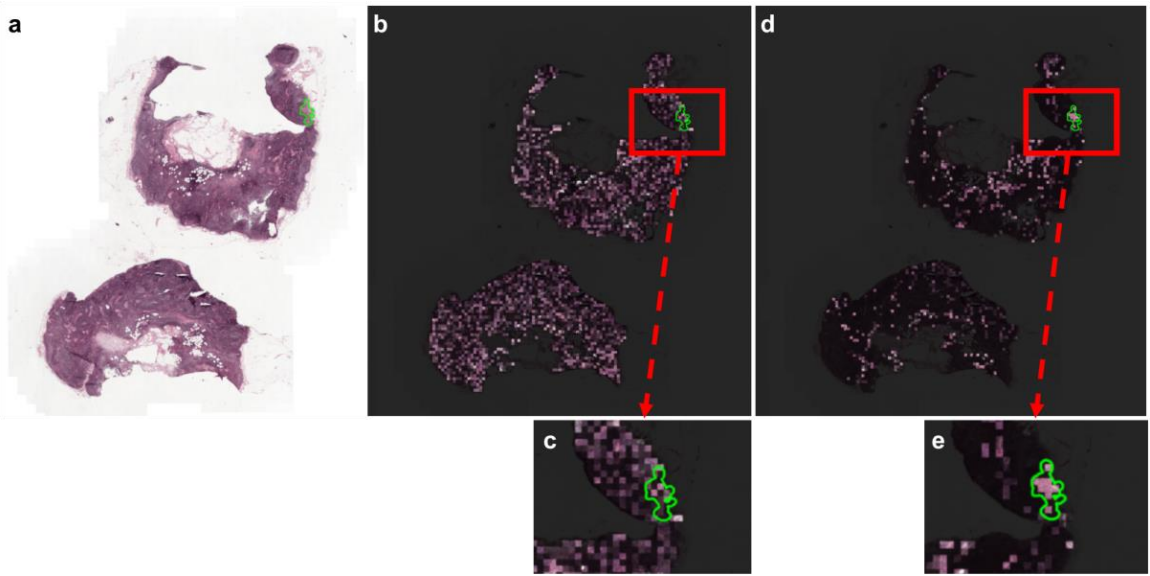

**Fig. 1.** Comparison of attention heatmaps for a tumor WSI produced by a current attention-based model and the proposed SiiMIL model. The heatmaps are visualized by overlaying the attention weights to the corresponding local regions of the WSI. Tumors are annotated in green. (a) Original tumor WSI with tumor annotation. (b) Attention heatmaps produced by an attention-based model. (c) Magnified local region. (d) Attention heatmaps produced by the proposed SiiMIL. (e) Magnified local region. From (b) and (c), we observe that the attention-based model fails to pay higher attention to the tumor region. In contrast, the proposed SiiMIL pays higher attention to the tumor region and pays relatively low attention to the non-tumor region.

Here, we propose a novel MIL model named salient instance inference MIL (SiiMIL) and its application to classification of breast cancer metastasis to lymph node, i.e., tumor vs. normal classification. In the proposed model, we first learn a set of representative normal patches from the normal WSIs. Then, we creatively infer the possible lesion patches of an input WSI by comparing the similarities between all the input patches and a set of representative normal patches. Finally, we aggregate the inferred patches and make a prediction using a deep-learning model. In our patch inferring method, we avoid training with noisy labels (patches that are predominant and do not correspond to the WSI label) because we do not rely on any tumor WSIs. To determine the salient instances, our model takes advantage of the fact that each patch in a normal WSI corresponds to the WSI label. So, it only uses patches from normal WSIs during the determination of salient instances. In other words, similar to open-world learning, our model learns to recognize normal representation while treating the tumor patches as an unknown class [25]. The main contributions of this paper are as follows:

• We present a novel method called Salient Instance Inference MIL (SiiMIL) to identify small-sized tumors from histopathology whole slide images.

• We also present the synergy of representation learning and weakly supervised learning in bringing interpretability and localization to the tumor detection process.

• We analyze the impact of tumor size on MIL models' accuracies, and show that the proposed model outperforms the existing models in terms of AUC, accuracy, sensitivity and interpretability.

• The proposed method has the potential to greatly reduce the need for IHC staining, thereby increasing efficiency in the overall clinical workflow.

## 2. Related works
### 2.1 Multiple instance learning in WSIs

Multiple instance learning (MIL) represents a weakly supervised learning approach in which labels are assigned to collections of instances, referred to as "bags", as opposed to the conventional machine learning practice of assigning labels to individual instances [26]. Typically, it performs binary classification on a bag by predicting the presence of "positive" instances. In 2018, Courtiol et al. conducted a pioneering study on predicting slide-level outcomes of Whole Slide Images (WSI) using MIL [19]. In their work, the focus was on predicting the presence of tumor metastasis in sentinel lymph nodes. Within this context, a slide was defined as a bag and patches containing tumor lesions were defined as positive instances. Their MIL approach addressed the GPU memory limitation issue in analyzing giga-pixel WSIs by partitioning WSI into small patches, and then embedding the patches into vectors. Following this work, multiple MIL methodologies were introduced for WSIs. These approaches can be categorized into two main groups: critical-instances-based approaches [8, 9, 19, 24, 27] and attention-based approaches [10, 20-22, 28]. Critical-instances-based methods usually predict slide-level outcome from a few selected patches with high confidence on-the-fly. For example, Campanella et al. [8] selected patches with top-$k$ highest probability predicted by their model and aggregated the selected patches using recurrence neural network, and Li et al. [9] predicted one critical patch and then aggregated other patches based on their similarity to the critical patch. On the other hand, attention-based approaches learn attention weights to highlight important patches during aggregation without selecting patches. Ilse et al. initially presented attention-based MIL (ABMIL) for natural images and tabular data. Lu et al. [22] extended the use of ABMIL to WSIs by adding clustering constraint. Zhang et al. [20] further partitioned a slide into multiple pseudo-bags of patches and performed ABMIL on each pseudo-bag separately. Other studies [10, 28] emphasized the importance of contextual relationship among the patches. As a result, these studies proposed to aggregate patches within a WSI using Transformer models [29]. Overall, it is challenging to accurately select critical-instances in weakly supervised learning, hence most of the recent studies utilize attention-based approach. However, the sensitivity of the existing MIL models are still moderate, especially for those WSIs with small lesions (i.e., corresponding to a few positive patches). In this study, we propose a salient instance inference-based MIL to alleviate this issue.

### 2.2 Deep learning in medical image analysis

In recent years, the application of deep learning has gained prominence in addressing similar demands for medical image analysis across various domains such as CT, X-ray, and MRI. Wen et al. introduced an attention capsule sampling network to enhance key slice in chest CT images for COVID-19 diagnosis [30]. Zhang et al. presented a double-branch attention network to perform tumor segmentation and metastasis classification simultaneously, in which two related tasks can share feature learning [31]. Leveraging CT imaging features for Osteoporotic Vertebral Fracture classification in X-rays, Wang et al. proposed a multi-modal neural network [32]. Besides, non-neural network methodologies like feature engineering and optimization algorithms have been explored in X-ray image segmentation tasks [33, 34]. Extending beyond radiology images, successful studies have emerged in clinical photographs and endoscopic images. For example, Su et al. devised an ensemble classifier to combine different neural network backbones for gastrointestinal disease from gastrointestinal endoscopic images [35]. Hu et al. proposed polyp extraction and reflection removal methods to highlight salient region in endoscopy images [36]. Exploring skin disease classification, Zhou et al. explore the impact of background color on convolutional neural networks in skin photograph analysis [37].

In contrast to these tasks, WSI classification poses a more challenging problem with exponentially higher image size and tiny portion of lesion areas, which makes common convolutional neural networks inadequate. Therefore, multiple instance learning has become the standard strategy for classifying WSIs.

## 3. Materials and methods

### 3.1 Dataset

Our study is developed based on the Camelyon16 dataset [18]. Camelyon16 is a publicly available hematoxylin and eosin (H&E) whole slide image (WSI) dataset for breast cancer lymph node metastases classification. The training set includes 271 WSIs, and the hold-out testing set has 129 WSIs divided into two classes: normal and tumor. In this weakly supervised learning study, only the WSI-level labels (i.e., normal or tumor) are available.

### 3.2 Multiple instance learning for WSI classification

In multiple instance learning (MIL) formulation, a bag refers to the set of unlabeled instances in a WSI. Each bag's label is determined by the presence of positive instances. In other words, a positive bag needs to contain at least one positive instance and a negative bag needs to exclusively contain negative instances. Therefore, we formulate the MIL problem as:

$$Y = \begin{cases} 0, if f \sum_i y_i = 0, \\ 1, \quad otherwise. \end{cases} \quad (1)$$

where $Y$ is the label of the bag, $y_i \in \{0, 1\}$, for $i = 1, ..., m$, is the label of instance inside the bag, and $m$ is the number of instances inside the bag [38]. To predict the bag-level label $Y$, the MIL models need to analyze and aggregate the instances, and then reach the conclusion in the form of:

$$Y = g(\sigma(f(x_1), ..., f(x_m))) \quad (2)$$

where $f(\cdot)$ is the feature extraction function for instances, $\sigma(\cdot)$ is the aggregation operator, and $g(\cdot)$ is the bag-level classifier.

For a variety of reasons, the WSI classification inevitably develops into the MIL formulation. First, the raw images of WSIs with gigapixel resolution seldom fit in the available graphics processor memory. As a result, the typical paradigm entails cropping a WSI into numerous image patches and then extracting feature embeddings from the patches using a feature extractor (i.e., the $f(\cdot)$ function). This enables us to fit all embedding into a GPU memory. Second, in WSI classification, such as normal/tumor, bags containing embeddings from lesion areas might be deemed positive because they are only present in tumor WSIs. Third, as it is painstakingly difficult to annotate WSIs, we often have access to the WSI-level labels (i.e., $Y$) but not the instance-level labels (i.e., $y_i$). The ability of MIL to address all three problems makes it a suitable candidate for WSI-level classification.

In the case of tumor WSI with small lesions, the WSIs typically contain dominant regions of normal tissues, making it difficult for the MIL model to focus on the patches corresponding to small lesions [22]. For instance, in Camelyon16 dataset, the ratio of tumor region to the entire tissue region of a tumor WSI could be lower than 0.5%.

### 3.3 Motivation of salient instance inference

To depict the motivation of our method, we take tumor WSIs classification as an example. Consider a tumor WSI (positive bag) with a small number of tumor patch feature embeddings (positive instances). Let $\{u_1, u_2, ..., u_p, v_1, v_2, ..., v_q\}$ denote $m = p + q$ instances of the WSI, where $u_i \in \mathbb{R}^D$ and $v_j \in \mathbb{R}^D$, for $i = 1, 2, ..., p$ and $j = 1, 2, ..., q$, denote the negative and positive instances, respectively. For the WSI with small lesions, we assume that $p \gg q$.

Assuming that our MIL aggregation function is weighted average:

$$z = \sum_{i=1}^{p} a_i u_i + \sum_{j=1}^{q} b_j v_j \quad (3)$$

where $z \in \mathbb{R}^D$ is the bag-level representation and $a_i$ and $b_j \in \mathbb{R}$ for $i = 1, 2, ..., p$ and $j = 1, 2, ..., q$ are the weights emphasizing the positive instances and $\sum_{i=1}^{p} a_i + \sum_{j=1}^{q} b_j = 1$. In other words, the bag-level representation is a convex combination of all negative and positive instances in a bag. Considering the morphological difference between tumor and normal tissues, the positive and negative instances should lay in different sub-spaces. However, since $p \gg q$, if the attention scores of the negative instances are not sufficiently small, the resulting convex combination $z$ will easily fall into the negative subspace. As a result, the positive bag-level representation will be inseparable from the negative bag. This effect is demonstrated in Section 3.1. On the other hand, if we intentionally select possible positive instances in the bag, therefore increasing the density of positive instances in the positive bags, we can make the bag-level representation more separable from the negative bag and derive an easier classification problem. Please see Fig. 2 as a toy example when the instance dimension $D = 2$.

For a naïve solution, if we can have an accurate instance-level positive instance prediction model, we can easily select high probability instances to achieve our goal. However, this type of prediction model usually requires fully supervised training with reliable instance labels, which is impossible for most WSI classification cases. Therefore, a weakly supervised patch selection methodology is crucial to tackling the MIL problem.

### 3.4 Salient instance inference

Our salient instance inference (Sii) is composed of the following three steps: (i) learning representative instances from negative (normal) WSIs in the training set; (ii) comparing similarities between the instances of an input WSI and the learned representative instances; (iii) selecting instances with low similarity scores (i.e., high saliency) from the input WSI to form a bag.

#### 3.4.1 Motivation for key instance selection through representation learning

Due to the anatomical redundancy in normal histopathological slides (normal WSIs or negative bags), our objective is to discern a subset of negative (or normal) patches (instances) that represent the essence of all normal patches within these WSIs. This subset is termed the *representative*

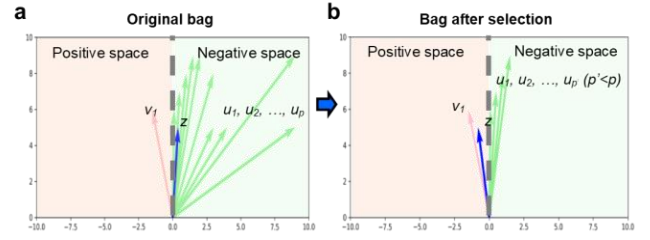

**Fig. 2.** Justification of salient instance inference in a simulated 2-D example. (a) We show a 2-D example of MIL aggregation for a small tumor WSI when all instances are used for aggregation. The red vector indicates the positive instance $v_1$, green vectors indicate the negative instances $u_{1,2,...p}$, and the blue vector indicate the combined bag-level representation $z$. (b) By selecting salient instances and omitting some easily recognizable negative instances, we reduce the number of negative instances from $p$ to $p'$, where $p' < p$. Thus, the bag-level representation $z$ will fall back to the positive space. In this toy example, all vectors are simulated, and horizontal/vertical axes indicated the two dimensions of the simulated 2D vectors.

*negative instances*. The primary motivation is to use these representative instances to pinpoint patches in an input WSI that most markedly deviate from normal negative instance features. In a normal WSI, patches inherently display patterns distinct from positive instances. Given the extensive collection of negative patches sourced from normal WSIs, we are equipped to reliably identify these representative negative instances.

To derive a set of representative negative instances, we first gather all negative patches from all normal WSIs. After processing them through a pre-trained feature encoder and subsequent concatenation, we produce a matrix with columns representing the feature embeddings of all normal patches. The task then is to determine an optimal subset that accurately reflects the wide-ranging distribution of normal patches. Inspired by CUR decomposition [39], renowned for its ability to refine a matrix by accentuating pivotal columns and rows based on their statistical significance or leverage, we employ a parallel strategy for our histopathological patches. Within the CUR paradigm, essential columns (or rows) are handpicked to comprehensively represent a matrix. In a similar approach, our methodology focuses on identifying the most representative embeddings associated with negative patches to ensure a robust representation of normal tissue patterns.

#### 3.4.2 Representation learning from negative instances

Let $X = [x_1, x_2, ..., x_n] \in \mathbb{R}^{D \times n}$ denotes a matrix of all patch embeddings inside a normal WSI where $x_i$ is the patch that is embedded by a feature extraction neural network.

Given a normal WSI with feature embeddings $X \in \mathbb{R}^{D \times n}$, if $n \gg D$, then there might exist $t \ll n$ feature embeddings that contain sufficient information to represent the WSI. To find out the $t$ most representative feature embeddings, we first apply the CUR decomposition to $X$ to construct a low-rank matrix approximation $\tilde{X} \in \mathbb{R}^{D \times t}$ as follows:

- Step 1. Compute the rank of $X$, and set $k = rank(X)$.
- Step 2. Construct matrix $V$ whose rows are the eigenvectors of $X^T X$.
- Step 3. Compute the importance score of the $j^{th}$ column of $X$ by

$$s_j = \frac{1}{k}\sum_{h=1}^{k} V_{hj}^2 \qquad (4)$$

where $V_{hj}$ is the element in the $h^{th}$ row and $j^{th}$ column of V, for $j = 1, 2, \ldots, n$.

- Step 4. Sort columns of $X$ based on the scores $s_j$'s.
- Step 5. Construct $\tilde{X} \in \mathbb{R}^{D \times t}$ whose columns are the first $t$ columns of sorted $X$ in Step 4.

We then apply this to all normal WSIs in our training set to obtain $\tilde{X}_i$, for $i = 1, 2, \ldots, P$, and concatenate to construct a key matrix $K$:

$$K = \tilde{X}_1 \oplus \tilde{X}_2 \ldots \oplus \tilde{X}_i = [k_1, \ldots, k_\tau] \in \mathbb{R}^{D \times \tau} \qquad (5)$$

where $\oplus$ denotes the concatenation operation and $\tau$ denotes the total number of representative negative instances from $P$ WSIs.

### 3.4.3 Bag generation

In this stage, for each WSI, we select salient instances to form a bag. Given all instances inside an input WSI, we think that the instances that are dissimilar to the representative negative instances (see 3.4.2) are with high saliency and should be included in the bag for the input WSI. In other words, we keep the instances that are more likely to be lesions and remove those simple-recognizable "normal" instances in the bag. As a result, we can yield a bag with a higher positive instance rate that is easily separable from the normal bag. We name this procedure salient instance inference (Sii). Please note that applying Sii on a normal WSI will not significantly affect the classification since the remaining instances of a negative bag will still be negative.

In detail, let $Q = [q_1, \ldots, q_n] \in \mathbb{R}^{D \times n}$ denotes the matrix of all instances of an input WSI where each column $q_i$ is an instance (i.e. patch embedding) of the WSI and $K = [k_1, \ldots, k_\tau] \in \mathbb{R}^{D \times \tau}$ denotes the representative negative instances from last stage. Inspired by the transformer model [29], we refer $q_i$ as queries and $k_j$ as keys. We build a cross similarity matrix $C \in \mathbb{R}^{m \times n}$ where each entry is the cosine similarity of a $(q_i, k_j)$ pair:

$$C_{ij} = \frac{q_i^T k_j}{\|q_i\|\|k_j\|}, \qquad i = 1, \ldots, n, \qquad j = 1, \ldots, \tau \qquad (6)$$

Then, for each query $q_i$, we define its saliency score $s_i \in \mathbb{R}$ as:

$$s_i = -Avg\left(\underset{j=1,\ldots,\tau}{TopK}(C_{ij})\right) \qquad (7)$$

where $TopK$ operator denote finding the top-$K$ highest $C_{ij}$ along the $j$ axis of the matrix $C$. Specifically, we are finding the $K$ nearest neighbors from all the keys for each query and then take the average of the corresponding similarity values. Since we define low similarity values as high saliencies, we take the negative average operation to produce the final saliency score $s_i$ for $q_i$.

Finally, we infer the salient query instances by selecting the queries with top-$r\%$ saliency score, where $r \in [0, 1]$ is a percentage number and form the bag $B = \{q_1, \ldots, q_t\}$, where $t = r \cdot n$, for this WSI using these query instances.

The processing steps of the completed Sii is depicted in Algorithm 1.

---

**Algorithm 1** Salient instance inference (Sii)

**Part 1.** Representation learning from negative instances:
**Input:** The set of normal WSIs, $X = \{X_1, X_2, \ldots, X_P\}$.
**Step 1:** for $i = 1, 2, \ldots, P$ do

- **Step 1.1:** Compute the rank of $X_i$, and set $k = rank(X_i)$.
- **Step 1.2:** Construct matrix $V$ whose rows are the eigenvector of $X_i^T X_i$.
- **Step 1.3:** Compute the importance score of the $j^{th}$ column of $X$ by $s_j = \frac{1}{k}\sum_{h=1}^{k} V_{hj}^2$, where $V_{hj}$ is the element in the $h^{th}$ row and $j^{th}$ column of $V$, for $j = 1, 2, \ldots, n_i$.
- **Step 1.4:** Sort columns of $X_i$ based on the scores $s_j$'s.
- **Step 1.5:** Construct $\tilde{X}_i \in \mathbb{R}^{D \times t_i}$ whose columns are the first $t_i$ columns of sorted $X_i$ in Step 1.4.

**End (for).**
**Output:** Construct a key matrix $K$ by concatenating all $\tilde{X}_i$, for $i = 1, 2, \ldots, P$,
$$K = \tilde{X}_1 \oplus \tilde{X}_2 \ldots \oplus \tilde{X}_i = [k_1, \ldots, k_\tau] \in \mathbb{R}^{D \times \tau}$$

**Part 2.** Bag generation:
**Input:**
- The keys matrix $K$ obtained in **Part 1.**
- All patch embeddings $Q = [q_1, \ldots, q_n]$ from a WSI (normal or tumor).

**Step 1:** Compute $C \in \mathbb{R}^{m \times n}$, where each entry is the cosine similarity of a $(q_i, k_j)$ pair.
**Step 2:** Compute $s_i = -Avg(TopK(c_{ij})_{j=1,2,\ldots,\tau})$ and $c_{ij}$ is the component of $C$.
**Step 3:** Compute $Q_{sort} = Sort([q_1, q_2, \ldots, q_n] | [s_1, s_2, \ldots, s_n])$, where $Sort(A | B)$ is a function that sorts the components of A based on the scores in $B$.
**Output:** Bag = $Top$-$r\%(Q_{Sort})$

---

### 3.5 MIL classification

Given the salient bag derived from the Sii process, we will aggregate the bag using MIL aggregation and finally make WSI-level prediction. The MIL aggregation function we are using is the attention-based MIL (ABMIL) [21]. Given a salient bag $B = \{q_1, \ldots, q_t\}$ for a WSI, we have the formulation:

$$z = \sum_{i}^{t} a_i q_i \qquad (8)$$

where:

$$a_i = \frac{exp\left(W^T(tanh(V^T q_i) \odot sigm(U^T q_i))\right)}{\sum_{j=1}^{o} exp\left(W^T(tanh(V^T q_j) \odot sigm(U^T q_j))\right)} \qquad (9)$$

where $V \in \mathbb{R}^{D \times L}$, $U \in \mathbb{R}^{D \times L}$ and $W \in \mathbb{R}^{L \times D}$ are the learnable weights of fully connection networks. Each instance $q_i \in \mathbb{R}^{D \times 1}$ is scaled by attention weight $a_i \in \mathbb{R}$ and summed to yield the representation vector $z \in \mathbb{R}^{D \times 1}$ for bag $B$. Finally, we apply a separate fully connection network on to $z$ to make WSI-level prediction. The overview of the complete SiiMIL is shown in Fig. 3.

### 3.6 Implementation details and evaluation metrics

In data preprocessing stage, we cropped each WSI into $224 \times 224$ image patches without overlapping under $20 \times$ magnification. We applied color thresholding method to extract foreground tissue patches and discarded background patches that contained less than 75% foreground tissue. Then, we embedded each patch with a ResNet50 model [40] pre-trained on ImageNet dataset [41]. The ResNet50 was truncated after the third residual block which embedded patches into 1024-dimension feature embeddings.

During the training of our MIL model, we used Adam Optimizer [42] with 0.0002 learning rate and 0.00001 weight decay. To avoid overfitting, we applied early stopping strategy with the patience of 10 epochs and the maximum training of 100 epochs depending on the validation loss. In other words, the training was halted if the validation loss do not decrease for 10 epochs.

We conducted five-fold Monte Carlo cross-validation on the training set with random splits in the ratio of training: validation set = 90:10. Then, we selected the model with the best validation AUC and evaluated it on the official testing set of Camelyon16. Our model contains three primary hyperparameters: $t_i$, $K$ and $r$. To clarify, $t_i$ controls how many representative instances that we select from each normal WSI using CUR-decomposition, $K$ controls the $topK$ operator during the cosine similarity comparison, and $r$ controls the top-$r$% salient instances selection. We tuned and selected the hyperparameters based on the validation AUC and decided to choose $t_i = 100$, $K = 150$, and $r = 0.3$.

We used five evaluation metrics to evaluate the WSI-level classification performance, which are accuracy, AUC (of ROC), precision, recall, and F1-score. They are defined as follows:

$$AUC = \int_0^1 TPR(t) \, dFPR(t)$$

where $TPR$ stands for true positive rate, $FPR$ stands for false positive rate, and $t$ indicates classification threshold changing from 0 to 1.

$$Accuracy = \frac{TP + TN}{TP + FP + TN + FN}$$
$$Precision = \frac{TP}{TP + FP}$$
$$Recall = \frac{TP}{TP + FN}$$
$$F1 \, score = \frac{2TP}{2TP + FP + FN}$$

where TP, TN, FP, and FN stand for true positives, true negatives, false positives, and false negatives.

Moreover, to evaluate the effect of Sii module, we defined the metric "tumor instance rate" (TIR) as:

$$TIR = \frac{Number \, of \, tumor \, instances}{Numer \, of \, all \, instances} \quad (8)$$

for a bag of instances.

Our code is available at: https://github.com/JoeSu666/SiiMIL.

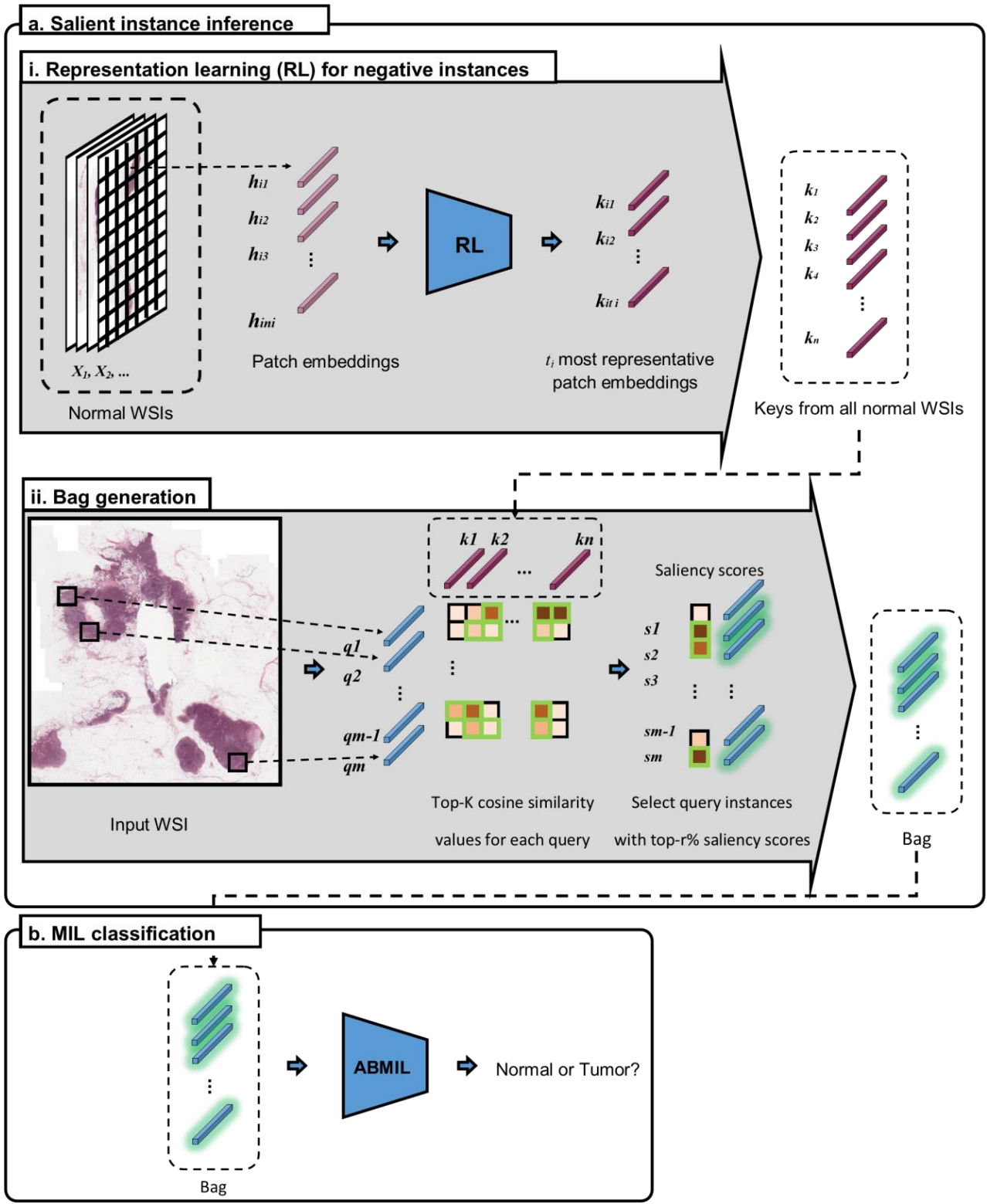

**Fig. 3.** Overview of our SiiMIL. Our model is comprised of two main steps: (a) salient instance inference, and (b) MIL classification. For the salient instance inference, (a.i) we first learn the representative normal patch embeddings, which we call keys, from all normal WSIs. (a.ii) Then, given an input WSI, we compare every instance with the keys, and select a bag of salient instances as the representation for the given WSI. (b) Finally, we use attention-based MIL (ABMIL) to predict the bag.

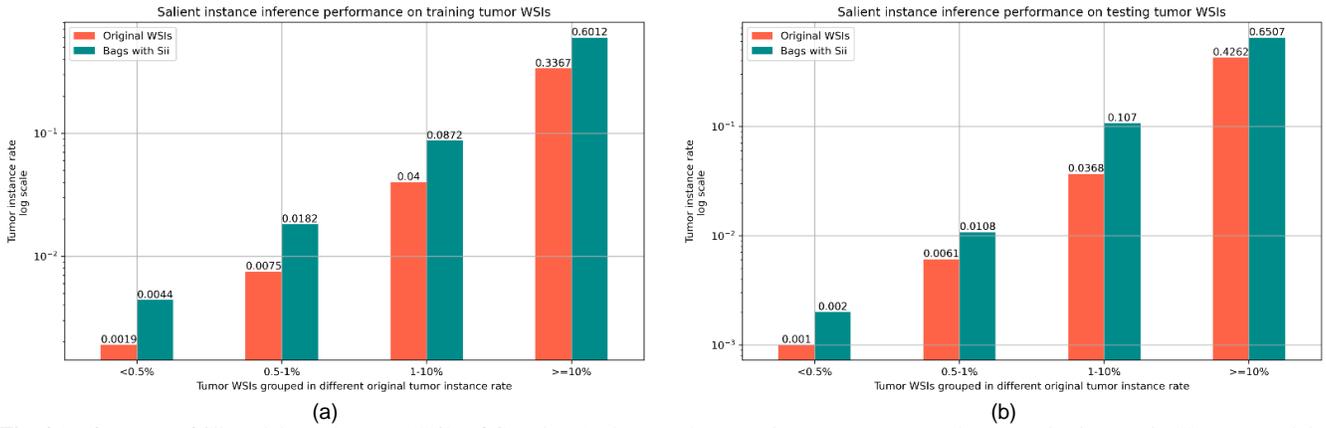

**Fig. 4.** Performance of Sii module on the tumor WSIs of Camelyon16 dataset. The tumor instance rate was used as an evaluation metric. We compared the TIR of original tumor WSIs, where all the instances in a WSI form a bag, with the TIR of corresponding WSIs after Sii. The WSIs are divided into different groups according to their original TIRs. (a) Sii performance on training tumor WSIs. (b) Sii performance on testing tumor WSIs.

**Table 1**

Classification results on Camelyon16 dataset with 95% CI [ll, ul]

| Methods | AUC | Accuracy | Precision | Recall | F1-score |
| --- | --- | --- | --- | --- | --- |
| ABMIL [21] | 0.8375 | 0.8372 | 0.8684 | 0.6734 | 0.7586 |
|  | [0.8349, 0.8419] | [0.8353, 0.8404] | [0.8668, 0.8759] | [0.6695, 0.6802] | [0.7530, 0.7612] |
| CLAM [22] | 0.8319 | 0.8295 | 0.8462 | 0.6735 | 0.7500 |
|  | [0.8298, 0.8367] | [0.8274, 0.8326] | [0.8380, 0.8475] | [0.6682, 0.6791] | [0.7440, 0.7524] |
| DSMIL [9] | 0.8265 | 0.7907 | **1.0000** | 0.4490 | 0.6197 |
|  | [0.8219, 0.8281] | [0.7886, 0.7939] | **[1.0000, 1.0000]** | [0.4429, 0.4540] | [0.6086, 0.6193] |
| TransMIL [10] | 0.8926 | 0.8759 | 0.9714 | 0.6938 | 0.8095 |
|  | [0.8889, 0.8939] | [0.8723, 0.8768] | [0.9693, 0.9739] | [0.6862, 0.6967] | [0.8021, 0.8098] |
| DTFD-MIL [20] | 0.8452 | 0.8217 | 0.7826 | 0.7347 | 0.7579 |
|  | [0.8450, 0.8514] | [0.8206, 0.8259] | [0.7791, 0.7889] | [0.7329, 0.7430] | [0.7530, 0.7608] |
| SiiMIL | **0.9225** | **0.8915** | 0.9487 | **0.7551** | **0.8409** |
|  | **[0.9202, 0.9243]** | **[0.8895, 0.8938]** | [0.9463, 0.9518] | **[0.7497, 0.7596]** | **[0.8349, 0.8417]** |

*Note. LL and UL represent the lower-limit and upper-limit of the 95% CI.*

## 4. Results

In this section, we demonstrate the experimental results of our method along with the comparison with existing methods. In addition, we exhibit the ablation study results and interpretability of our SiiMIL.

### 4.1 Results of tumor instance inference

To validate the effectiveness of Salient instance embeddings inference, we grouped the WSIs based on their TIR in four groups: (i) < 0.5%, (ii) 0.5-1%, (iii) 1-10%, (iv) >=10%. In Fig. 4, we exhibited the performance of salient instance inference across different groups by comparing the tumor instance rate (TIR) of original tumor WSIs and the tumor bags after applying Sii.

### 4.2 Results on WSI classification

In Table 1, we presented our performance on Camelyon16 dataset and compared them with a set of state-of-the-arts MIL-based WSI classification models [9, 10, 20-22]. The proposed SiiMIL achieved 0.9225 AUC which outperforms the comparison models. To be noticed, the ABMIL was the baseline method of SiiMIL that didn't perform Sii and applied the attention module to all the instances of the WSIs. We observed more than 8 points improvement in AUC from the baseline to our SiiMIL.

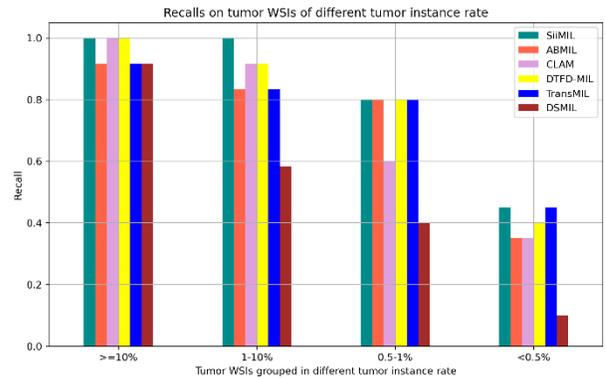

**Fig. 5.** Classification recall on tumor WSIs grouped in different TIRs

Due to the remarkable bag TIR improvement by using the Sii, our SiiMIL also achieved the best recall compared to the comparison methods, which is crucial in medical domain. To demonstrate the effect of tumor instance rate on recall, we grouped the tumor WSIs in different TIRs: (i) < 0.5%, (ii) 0.5-1%, (iii) 1-10%, (iv) >=10%. Then, we evaluated the MIL models' recall in every group of tumor WSIs (see Fig. 5). Here, we observed that our SiiMIL achieved outstanding accuracy in all of the groups.

### 4.3 Salient instance inference as an addition to MIL models

Other than an integrated MIL model, our Sii module can be also treated as a powerful addition to other MIL models. In Table 2, we presented the performance of some MIL models that used ResNet50 feature extractor as our model did. In parallel, we presented their performance where we applied Sii to form bags before MIL process. We observed that, by feeding salient bags yielded by our Sii, most of the MIL methods (i.e., ABMIL, CLAM, and DTFD-MIL) achieved improved performance. These methods, like ours, assume the permutation invariance property among instances, making them resilient to our instance selection strategy. Conversely, TransMIL, which leverages a Transformer architecture to capture contextual information, experienced a decline in performance. This outcome is reasonable, as our instance selection strategy potentially disrupts the contextual relationships within bags.

**Table 2**

Performance of Sii as an addition to other MIL models with 95% CI

| Methods | AUC | ACCURACY |
|---|---|---|
| ABMIL [21] | 0.8375 [0.8349, 0.8419] | 0.8372 [0.8353, 0.8404] |
| Sii+ABMIL | **0.9225** [**0.9202, 0.9243**] | **0.8915** [**0.8895, 0.8938**] |
| CLAM [22] | 0.8319 [0.8298, 0.8367] | 0.8295 [0.8274, 0.8326] |
| Sii+CLAM | **0.8747** [**0.8711, 0.8772**] | **0.8527** [**0.8494, 0.8543**] |
| DTFD-MIL | 0.8452 [0.8416, 0.8480] | 0.8217 [0.8175, 0.8229] |
| Sii+DTFD-MIL | **0.8630** [**0.8592, 0.8655**] | **0.8527** [**0.8509, 0.8558**] |
| TransMIL | **0.8926** [**0.8889, 0.8939**] | **0.8759** [**0.8723, 0.8768**] |
| Sii+TransMIL | 0.8676 [0.8672, 0.8731] | 0.8450 [0.8431, 0.8481] |

### 4.4 Ablations studies

We further conducted ablation studies to investigate the effects of three primary hyperparameters to our model: $t_i$, $k$, and $r$. To investigate the effect of numbers of CUR-based keys, we created key sets by setting $t_i = 50, 100, 200, 300$ and $400$, and named the derived key sets as: (i) CUR-50, (ii) CUR-100, (iii) CUR-200, (iv) CUR-300, and (v) CUR-400. Then, we applied those key sets separately in our SiiMIL and evaluated the resulted mean validation AUCs based on a five-fold cross-validation. From Fig. 6a, we found that our method achieved best performance when we used the CUR-100 key sets.

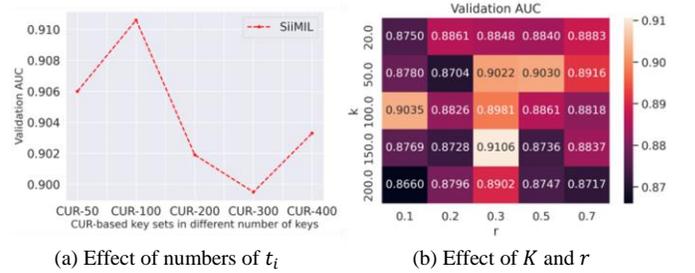

(a) Effect of numbers of $t_i$      (b) Effect of $K$ and $r$

**Fig. 6.** Ablation study on hyperparameters. (a) Effect of numbers of CUR-based keys. (b) Effect of $K$ and $r$. All metrics in the figures were the averaged validation AUCs based on the five-fold cross-validation results.

In Fig. 6b, we presented the mean validation AUCs of our method when we used different $(k, r)$ settings. We observed that our method achieved the best performance when we chose $k = 150$ and $r = 0.3$.

### 4.5 Interpretability of SiiMIL

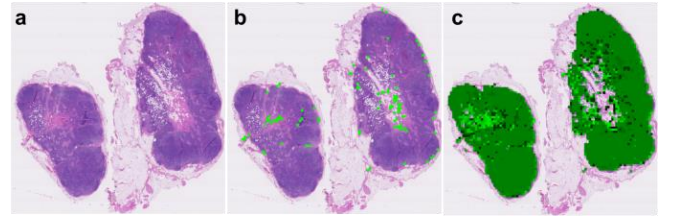

**Fig. 7.** Visualization of keys from a normal WSI. a) An example of a normal WSI. (b) The placement of the keys on top of the normal WSI. Each green block represents one of the 100 keys learned from the WSI. (c) Patch colormap, where patches of the same color share the same key representation.

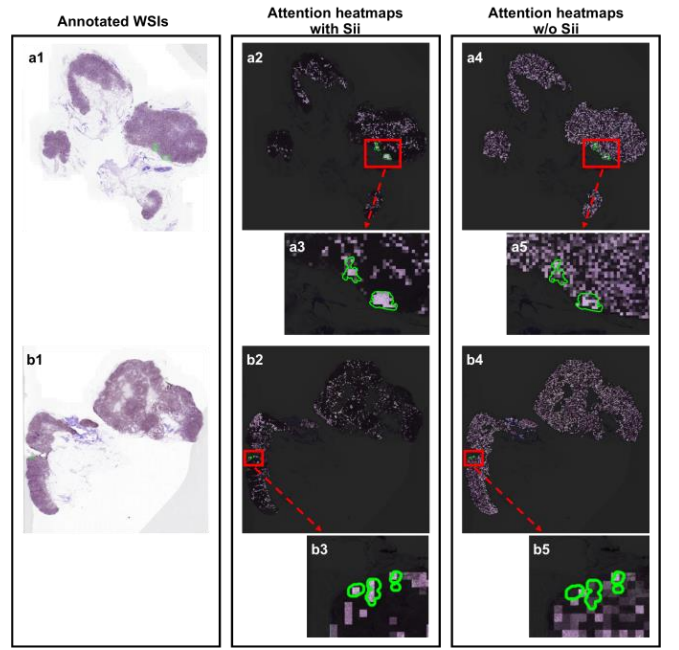

**Fig. 8.** Attention heatmaps visualization for example small tumor WSIs. The middle column shows the heatmaps produced by our SiiMIL. As comparison, the right column shows the heatmaps produced by the MIL model without using Sii. The attention weight of each patch is assigned by the attention module of the MIL models. Here, we visualize the weights onto the corresponding patches of the WSIs, so that the brighter regions are corresponding to the instances with high attention weights. The attention weights are scaled within 0-1 to enhance the contrast for visibility purpose.

Fig. 7 depicts a normal WSI as well as the position of the representative negative instances (keys). Fig. 7a depicts a typical WSI from our dataset. The location of the 100 keys selected by the proposed representation learning methodology are depicted in Fig. 7b. It is clear that these 100 keys correspond to different anatomical structures within the example WSI. Fig. 7c depicts a heatmap in which each instance embedding is colored the same as its most similar key (measured using cosine similarity in the feature space).

Our SiiMIL is also of high interpretability which is important for a diagnostic tool. To demonstrate the interpretability of SiiMIL, we visualized some small tumor WSIs' attention heatmaps that produced by the attention module of our model in Fig. 8 (middle column). As comparison, we also visualize the corresponding attention heatmaps produced by the MIL model without using our Sii (i.e., original attention-based MIL) in Fig. 8 (right column). In the heatmaps, the regions with high brightness correspond to the instances that received high attention weights. We can clearly observe that the SiiMIL is sensitive to the tumor regions during predictions, even on the small tumor WSIs.

## 5. Discussion

In this study, we develop a novel salient instance inference-based multiple instance learning algorithm for whole slide image (WSI) classification. Our contributions are twofold: (i) we develop a CUR-based representation learning method to mine the representative negative instances from negative bags (i.e., normal WSIs); (ii) we develop a salient instance inference method that can effectively select salient instances (i.e., possible positive instances) from a WSI so that we can form a bag by collecting salient instances only and feed our MIL model with this easy-to-classify bag.

The primary concept behind our method is to identify and feed MIL model with only the salient instances, simplifying the downstream classification problem. To achieve this goal, we propose salient instance inference (Sii) where we innovatively infer salient instances by comparing similarities between incoming instances and the representative negative instances. This idea is inspired by the open-set learning [43, 44] and anomaly detection [25, 45] technologies. In open-set learning and anomaly detection, we start by learning from known classes and then weed out data from unknown ones. In our problem formulation, the 'seen class' corresponds to normal WSIs, while tumor slides fall under the 'unseen class' category. This approach suits WSI classification because hospitals have many normal WSIs, easily labeled as such without annotations, while tumor slides are an unknown class that we aim to exclude. Therefore, in our method, we first conduct a representation learning on the instances from the normal WSIs using CUR decomposition. To the best of our knowledge, this is the first study of using CUR to learn representations for the WSIs. After learning a strong set of "key" negative instances, given a query WSI, we propose a transformer-like architecture [29, 46] to compare cosine similarities between the query instances and the keys. Leveraging the difference between negative and positive patterns, we hence infer the most dissimilar query instances and claim them as salient instances.

By applying our Sii module to the tumor WSIs, we effectively improve the tumor instance rates (TIRs) of the resulting bags two to four times compared to the TIRs of the original WSIs. Notably, we achieve this level of improvement even on the tumor WSIs whose original TIRs are lower than 1%, which are extremely difficult cases for the WSI classification problem. Based on the effectiveness of Sii, our SiiMIL achieves an AUC of 0.9225 (95% CI: [0.9202, 0.9243]) and an accuracy of 0.8915 (95% CI: [0.8895, 0.8938]), which outperforms the state-of-the-art MIL models. In addition, we show that existing MIL models' performance in terms of recall will decrease with decreasing TIRs of the WSIs (see Fig. 5). This observation verifies our motivation of improving the positive instance rate (see Section 3.2 for details). Despite declining TIRs, SiiMIL still outperforms existing MIL models in recall, further supporting our idea to raise the positive instance rate. Sii is also a powerful addition to the existing MIL models. In Table 2, we exhibit that Sii improves the AUCs of three existing MIL models (ABMIL [21], CLAM [22], and DTFD-MIL [20]) by 3-9 points. These three models all assume permutation invariance property among the instances. In contrast, TransMIL, which employs Transformer to capture contextual relationship between instances, experienced degraded performance after applying our Sii method. This is understandable, as our instance selection strategy could potentially disrupt the contextual relationships within the bags. Lastly, our SiiMIL method has excellent interpretability. From the attention heatmaps shown in Fig. 8, we can clearly observe that our model accurately pays attention to the tumor regions during predictions, even on the small tumor WSIs.

## 6. Conclusions and future works

Here, we propose a salient instance inference based multiple instance learning (SiiMIL) model that can accurately classify whole slide images (WSIs). A distinctive aspect of our work is addressing the impact of small lesion sizes on MIL model performance within the context of WSI tasks. To overcome this challenge, we have introduced a novel representation learning technique for histopathology images, which identifies representative normal keys. These keys enable the selection of salient instances within WSIs through our salient instance inference algorithm, resulting in the creation of bags with high tumor instance ratios (TIRs). Ultimately, the incorporation of an attention mechanism facilitates slide-level classification using the constructed bags.

The main limitation of our method lies in the fact that it is tailored specifically for binary classification problems. In our future studies, we will generalize it to multi-class classification problems. Our method can also be improved by taking the contextual information into account in the MIL aggregation stage. However, since we omit 70% of the instances using our Sii, we cannot simply apply our method to the existing context-based MIL models. Instead, we can develop special position encoding methods for our Sii to fit those models as part of our future work. Moreover, we are planning to make our Sii module trainable so that we can simplify the hyperparameters and learn better salient instances constrained by the classification loss function.

In spite of the limitations, our SiiMIL is an innovative and effective model for the WSI classification task. In addition, our Sii method creates a new paradigm that different negative instance representation learning method can be explored and applied to our method for better salient instance inference outcomes. Moreover, we believe that our idea about instance selection for MIL models will trigger more discussion on how to improve the performance of MIL models by selecting instances. Finally, the proposed Sii method can be also applied in the WSI image retrieval task, which is a hot topic recently [48, 49]. Briefly, our CUR-based representation learning can be used as a database generation method, and our similarity comparison strategy can be used as the image retrieval method in the WSI image retrieval context.

From clinical perspective, the proposed model exhibits outstanding accuracy and interpretability, even for the WSIs with extremely small tumor regions. This property could enhance pathologists' sensitivity in diagnosing based on H&E slides, leading to improved early detection of tumors and ultimately positively impacting patients' lives. Additionally, it may contribute to a reduction in the necessity for additional IHC staining in clinical practice. We believe that our method could yield multiple successful applications in different medical scenarios and therefore improve the clinical diagnosis.

**Declaration of Competing Interest**

The authors declared that they have no conflicts of interest in this work.

**Acknowledgement**

The research was partially funded by a National Institutes of Health R01CA276301 (PIs: Niazi, Wei), Trailblazer award R21EB029493 (PIs: Niazi, Segal), R21CA273665 (PI: Gurcan), R01DC020715 (PIs: Gurcan and Moberly), and Alliance Clinical Trials in Oncology GR125886 (PIs: Frankel and Niazi). The content is solely the responsibility of the authors and does not necessarily represent the official views of the National Institutes of Health.